%% file: ICCVW.tex
\documentclass[10pt,twocolumn,letterpaper]{article}

\usepackage{iccv}
\usepackage{times}
\usepackage{epsfig}
\usepackage{graphicx}
\usepackage{amsmath}
\usepackage{amssymb}

\usepackage{xcolor}

\usepackage{graphicx}
\usepackage{amsmath}
\usepackage{amssymb}
\usepackage{subcaption}
\usepackage{color,soul}
\usepackage{multirow}
\usepackage{enumitem}

\newcommand*{\ICCVW}{}

\usepackage[pagebackref=true,breaklinks=true,letterpaper=true,colorlinks,bookmarks=false]{hyperref}

\iccvfinalcopy 

\def\iccvPaperID{0006} 

\ificcvfinal\fi

\begin{document}

\title{FisheyeMODNet: Moving Object detection on Surround-view Cameras for Autonomous Driving}


\author{Marie Yahiaoui$^1$$^,$$^2$, Hazem Rashed$^3$, Letizia Mariotti$^1$, Ganesh Sistu$^1$, \\ Ian Clancy$^1$, Lucie Yahiaoui$^1$, Varun Ravi Kumar$^4$ and Senthil Yogamani$^1$\\
{\normalsize $^1$Valeo Vision Systems, Ireland \hspace{0.3cm} $^2$ECE Paris, France \hspace{0.3cm}  $^3$Valeo R\&D, Egypt \hspace{0.3cm} $^4$Valeo DAR Kronach, Germany} \\ 
{\tt\small firstname.lastname@valeo.com}
}

\maketitle
\ificcvfinal\thispagestyle{empty}\fi

\input{Core/abstract.tex}

\section{Introduction}
\input{Core/intro}

\section{Moving Object Detection}
\label{sec:related}
\input{Core/related}

\section{Dataset Creation}
\label{sec:method}
\input{Core/method}

\section{Proposed Model and Experiments}
\label{sec:exps}
\input{Core/exps}

\section{Conclusion}
\label{sec:conc}
\input{Core/conc}

{\small
\bibliographystyle{ieee}
\bibliography{IEEEfull}
}

\end{document}

%% file: Core/abstract.tex
\begin{abstract}
Moving Object Detection (MOD) is an important task for achieving robust autonomous driving. An autonomous vehicle has to estimate collision risk with other interacting objects in the environment and calculate an optional trajectory. Collision risk is typically higher for moving objects than static ones due to the need to estimate the future states and poses of the objects for decision making. This is particularly important for near-range objects around the vehicle which are typically detected by a fisheye surround-view system that captures a 360$^\circ$ view of the scene. In this work, we propose a CNN architecture for moving object detection using fisheye images that were captured in autonomous driving environment. As motion geometry is highly non-linear and unique for fisheye cameras, we will make an improved version of the current dataset public to encourage further research. To target embedded deployment, we design a lightweight encoder sharing weights across sequential images. The proposed network runs at  15 fps on a 1 teraflops automotive embedded system at accuracy of 40\% IoU and 69.5\% mIoU.
\end{abstract}

\ifdefined\ICCVW

\else

\textbf{Keywords:} Automated Driving, Visual Perception, Fisheye cameras, Moving object detection.

\fi

%% file: Core/intro.tex
Vision based driver assistance systems \cite{horgan2015vision} have become very common in commercial vehicles and they are gradually moving towards higher levels of autonomous driving. Deep learning,  semantic segmentation in particular \cite{siam2017deep}, has played a significant role in enabling this progress.
Deep learning algorithms are computationally intensive and it is necessary to design efficient algorithms \cite{siam2018rtseg, briot2018analysis} for deployment. Deep learning algorithms also have the advantage of being used for various tasks by sharing the encoded features \cite{sistu2019neurall, chennupati2019multinet++}.
The autonomous driving scenes are highly dynamic, where there are a lot of moving objects interacting with each other forming a very complex environment to deal with. Knowing the motion information helps generic foreground detection \cite{jain2017fusionseg} and improves semantic segmentation \cite{rashed2019motion}. Fewer classes are movable and this can be leveraged to improve the classification accuracy. For example, object classes like buildings or poles are static and will not have dominant motion vectors after egomotion compensation.
There are two types of motion in an autonomous driving scene. The first one is motion of the surrounding obstacles and the second is motion of the ego-vehicle. The ego-motion might cause difficulties to successfully detect the moving objects because even static objects will be perceived as moving. Motion segmentation implies two tasks that are performed jointly. The first one is object detection in which we highlight the interesting objects only of specific classes, which are pedestrians and vehicles, and discard any motion perceived from the background due to ego-motion. The second is motion classification in which a binary classifier predicts whether the object is moving or static.  \ifdefined\ICCVW \\ \fi

\ifdefined\ICCVW

\begin{figure}[t]
\centering
\includegraphics[width=\columnwidth]{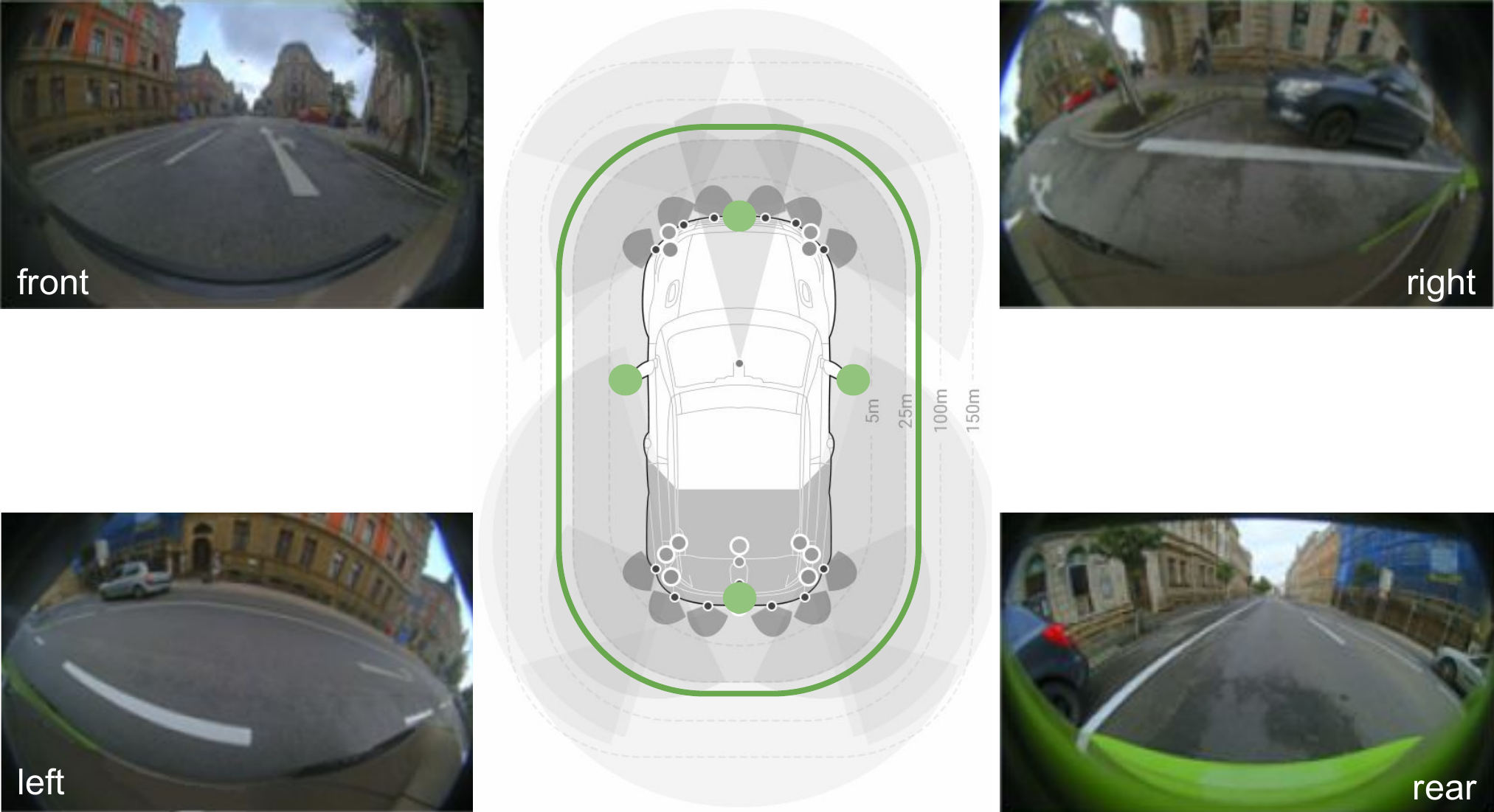}
\caption{Images from the surround-view camera network showing near field sensing and wide field of view. Four fisheye cameras (marked green) provide 360$^\circ$ surround view.}
\label{fig:svs}
\end{figure}

\fi

\ifdefined\ICCVW
\begin{figure*}[htb]
\centering
\includegraphics[width=\textwidth]{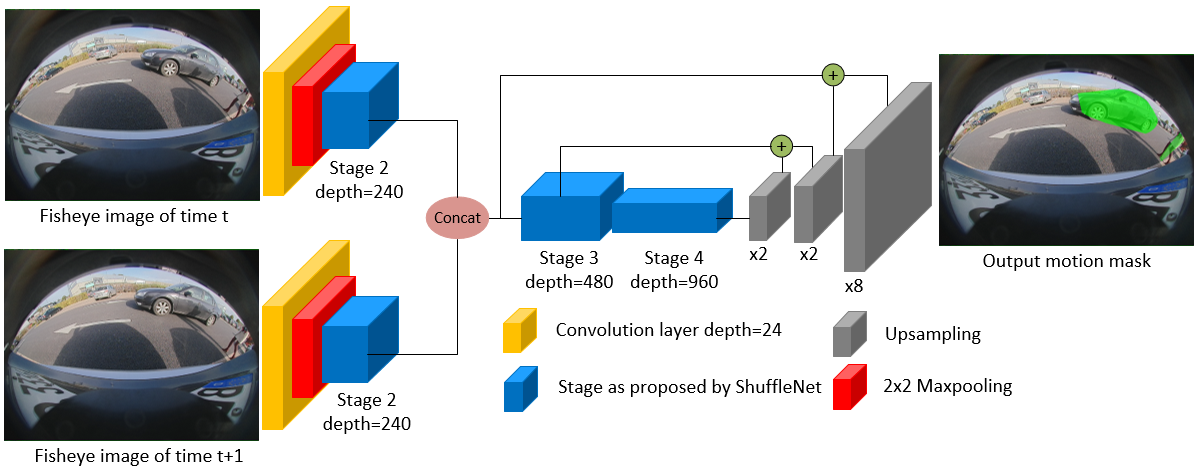}
\caption{Network Architecture adapted from ShuffleSeg base network. Two sequential images encoding the motion information across time are utilized train the network end-to-end for MOD.}
\label{fig:networkArchitecture}
\end{figure*}
\fi

In this work, we collect 5k samples of fisheye images captured using real cameras embedded on a moving vehicle in a 360$^\circ$-surround-view setup \ifdefined\ICCVW
(illustrated in Figure \ref{fig:svs}) \fi. We generate the annotation using a semi-automatic approach in the form of binary masks highlighting moving obstacles. We make use of the generated annotation to train an adapted end-to-end network which is based on \cite{gamal2018shuffleseg} for moving object detection. The algorithm leverages a two-stream mid-fusion approach, however we make use of two sequential images which encode motion across time where the network implicitly learns to distinguish between ego-motion and obstacles motion for the final motion segmentation task. The contributions of our work can be listed as follows:
\begin{itemize}[nosep]
   \item Generation of the first public automotive dataset for fisheye images with MOD annotations.
   \item Implementation of an efficient two-stream network architecture suitable for embedded systems.
   \item Empirical study of different training and data augmentation schemes.
\end{itemize}

%% file: Core/related.tex

The detection and localization of moving obstacles is critically important for Advanced Driver Assistance Systems (ADAS) and autonomous vehicles as they are essential for emergency braking, to support decision making for its next step navigation and to avoid possible collisions \cite{heimberger2017}. 
In automotive scenarios, rear-view and surround-view fisheye cameras are commonly deployed in existing vehicles for viewing applications. From a static observation point, the detection of moving obstacles is almost trivial as any non-zero optical flow will be due to motion in the scene or noise in the image. For a moving observer, the problem is challenging as the entire scene relative to the camera is moving and is additionally complicated when we consider fisheye cameras, which exhibit complex patterns of motion due to the non-linear projection and strong lens distortion. \ifdefined\ICCVW \\ \fi

\textbf{Related work:}
The classical approach to the detection of moving objects is based on the geometrical understanding of the scene, where the ego-vehicle motion and the displacement vectors of the pixels between two frames are known. Arguably the most famous constraint used in motion detection is the epipolar constraint \cite{ soumya2012, clarke1996}, 
which can be combined with additional geometrical constraints in order to detect multiple types of motion \cite{klappstein2006}. 
However, even if the geometry of moving objects is well known, their detection still presents challenges caused by the intrinsic geometrical limitations. 
In the search to overcome the limitations of the classical approach there has been promising work in using CNN to solve the moving object detection problem, such as MODNet \cite{siam2017modnet} and MPNet \cite{wang2018}. 
Given the use of fisheye cameras in surround-view systems, it is of utmost importance for research to explore this direction and provide a CNN architecture for moving objects detection on fisheye images. One of the main challenges of detecting moving objects with a CNN is to make it scene agnostic, so that the detection is based only on motion cues \& not on appearance cues.

%% file: Core/method.tex
\textbf{Fisheye Cameras:} Fisheye cameras are commonly used for near-field sensing for use cases like parking and traffic jam assist. They provide a wide field of view and requires just four cameras for the full 360$^\circ$ coverage. This advantage comes with a cost that is significantly more complex projection geometry exhibited by fisheye cameras. Thus models learnt on rectilinear cameras do not generalize well to fisheye cameras. This motivated us to create a new dataset focused on parking scenes with other vehicles and pedestrians being the main moving objects. \ifdefined\ICCVW \\ \fi

\ifdefined\ICCVW
\begin{table*}[htb]
\vspace{-0.2cm}
\centering
\caption{Quantitative evaluation on our fisheye images from test set. }
\begin{tabular}{llll}
\hline
\textbf{Model} & \textbf{Number of samples} & \textbf{mIoU} & \textbf{MOD IoU} \\ \hline
Trained on rectilinear KITTI data & 1300	& 53.5	& 10\\ \hline
Trained on fisheye data & 3638 & 69.5	& 39.8\\ \hline
+ weight sharing in two stream  & 3638 & 69.5	& 39.6\\ \hline
+ static objects scene augmentation & 5849 & 70 & 42 \\ \hline
\end{tabular}
\vspace{-0.4cm}
\label{table:fisheyResults}
\end{table*}
\fi

\textbf{Semi-automated annotation procedure}
There is no public dataset for fisheye images that focuses on  autonomous driving scenes, thus we introduce our own dataset which has 1 Megapixel images captured at 30 fps.  In order to train our network end-to-end for moving object detection, we developed a pipeline that generates MOD annotation to be used as ground-truth as illustrated in the procedure below: 
\begin{itemize} [nosep]
    \item Previously generated object annotation bounding boxes are parsed to identify the objects positions within the scene.
    \item The object positions from the annotations are used to extract from the LiDAR data those points that are within the annotated object.
    \item The extracted point cloud is then processed to classify the object as moving or static.
    \item After processing, the points are projected onto the image using the camera calibration information.
    \item The resulting set of 2D points is converted to a convex hull polygon.
\end{itemize}

\textbf{\\ Dataset Statistics:} The fisheye dataset used was generated from only parking scenes and contains a total of 5139 frames using the sampling strategy discussed in \cite{DBLP:conf/visapp/UricarHKY19}. We split the data into 70\% for training and 30\% for testing. A total of 3638 frames were used to train the network including 73 different scenes and 1501 frames were used to test the network, including 70 different scenes. The total number of moving objects annotations in the training dataset is 6296. The average number of moving objects per frame is 1.4, mainly pedestrians and cars. The average percentage of moving pixels in a frame is 0.54\%, and the average percentage of static pixels is 99.46\%, including background and static objects. The dataset will be released as part of WoodScape project \cite{yogamani2019woodscape}.




%% file: Core/exps.tex
\textbf{Proposed Model:}
The architecture used is based on \cite{gamal2018shuffleseg}, where the network is adapted to accept two-inputs in a two-stream fashion as proposed by \cite{siam2017modnet,jain2017fusionseg,8594088}. However, those methods used optical flow images to capture motion information and RGB images to understand scene semantics. Optical flow requires preprocessing, especially for fisheye images which will be distorted depending on the fisheye camera parameters. In our approach, we train the network end-to-end using temporally sequential images which encode both semantics and motion together. The network encoder is responsible for the feature extraction phase before the feature maps are upsampled to the input image size. The encoder is based on \cite{8594088} which utilizes point-wise group convolutions and channel shuffling, which dramatically reduce computation cost at a high accuracy level. The decoder part is composed of three deconvolution layers which provide the final output image size. The main advantage of this approach is its low complexity where a lightweight architecture is used to fit on autonomous driving embedded platforms and provide good accuracy as well. The network is trained to classify the output pixels among two classes, moving and non-moving classes. The number of static pixels exceeds the number of moving pixels. This is because of the background pixels which are considered as static ones, in addition to the static foreground pixels such as static vehicles and pedestrians. Weighted cross-entropy is utilized to overcome the class imbalance problem. \ifdefined\ICCVW \\ \fi

\ifdefined\ICCVW
\else
\begin{figure*}[ht!]
\centering
\includegraphics[width=0.8\textwidth]{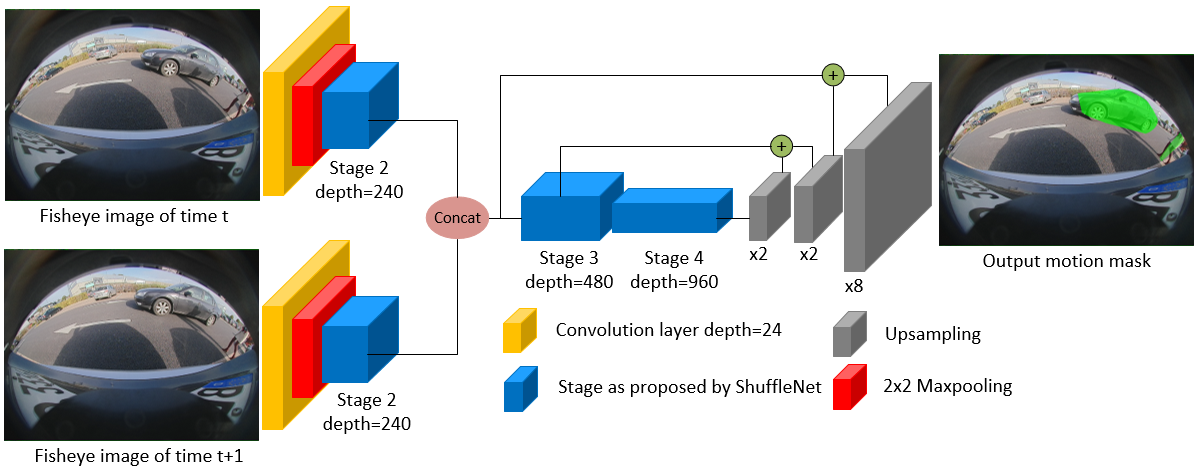}
\caption{Network Architecture adapted from ShuffleSeg base network. Two sequential images encoding the motion information across time are utilized train the network end-to-end for MOD.}
\label{fig:networkArchitecture}
\end{figure*}
\fi

\textbf{Experimental Setup:}
Figure \ref{fig:networkArchitecture} illustrates the network architecture we use where two temporally sequential images are processed separately in two encoders. This setup allows the network to understand the motion within the surrounding scene. The network is trained to generate a binary mask for MOD where each pixel can be moving or static. Throughout all experiments, weighted cross-entropy has been utilized to overcome the class imbalance problem. Adam optimizer is set at rate $1e^{-4}$. L2 regularization with weighted decay of $5e^{-4}$ has been utilized to avoid over-fitting the data. The network encoder is initialized with pre-trained weights on ImageNet. \ifdefined\ICCVW \\ \fi

\textbf{Results:}
Table \ref{table:fisheyResults} illustrates our results on MOD task using fisheye images. The first row represents the usage of pre-trained weights, where the network was trained on 1100 images using the dataset provided by \cite{siam2017modnet} and inference was done on fisheye images. Results show inability of the network to generalize rectilinear model to fisheye images. The second row shows results where the network is trained on 3k fisheye images where significant improvement was observed providing 40\% IoU compared to 10\% when trained on KittiMoSeg\cite{siam2017modnet} dataset. The third row shows further improvement after augmentation of the dataset with static objects so that the number of moving and static objects become balanced.
Overall, the detection results are reasonable and the main issue is with false positives with static pedestrians being detected as moving objects. This was due to small movement of pedestrians while standing still in the dataset. To improve efficiency, we used shared weights in the two encoders so the previous encoder can be re-used from the previous iteration. This resulted in very little decrease in accuracy as shown in fourth row. Finally, we augmented scenes which contained only static objects, which did not need any annotation. This resulted in a slight increase in accuracy as shown in the fifth row. The proposed network is very efficient and runs realtime at 15 fps on a 1 teraflops automotive embedded system.

\ifdefined\ICCVW
\else
\begin{table*}[ht!]
\vspace{-0.2cm}
\centering
\caption{Quantitative evaluation on our fisheye images from test set. }
\begin{tabular}{llll}
\hline
\textbf{Model} & \textbf{Number of samples} & \textbf{mIoU} & \textbf{MOD IoU} \\ \hline
Trained on rectilinear KITTI data & 1300	& 53.5	& 10\\ \hline
Trained on fisheye data & 3638 & 69.5	& 39.8\\ \hline
+ weight sharing in two stream  & 3638 & 69.5	& 39.6\\ \hline
+ static objects scene augmentation & 5849 & 70 & 42 \\ \hline
\end{tabular}
\vspace{-0.4cm}
\label{table:fisheyResults}
\end{table*}
\fi

 

%% file: Core/conc.tex
In this paper, we introduced a new moving object detection dataset for fisheye cameras. Firstly, we showed that the model trained on rectilinear KITTI dataset does not generalize well for fisheye images. We designed an efficient architecture for moving object segmentation and provided baseline experiments. We also tested different training and augmentation techniques to improve accuracy. We will make an improved version of the dataset public in order to encourage further research. In future work, we plan to incorporate geometric priors into the loss function to improve accuracy.